# Elimination of ISI Using Improved LMS Based Decision Feedback Equalizer

Mohammad Havaei, Nandivada Krishna Prasad, and Velleshala Sudheer

*Abstract*— This paper deals with the implementation of Least Mean Square (LMS) algorithm in Decision Feedback Equalizer (DFE) for removal of Inter Symbol Interference (ISI) at the receiver. The channel disrupts the transmitted signal by spreading it in time. Although, the LMS algorithm is robust and reliable, it is slow in convergence. In order to increase the speed of convergence, modifications have been made in the algorithm where the weights get updated depending on the severity of disturbance.

## I. INTRODUCTION

THE field of digital communication is increasing day by day. There have been vast changes going on. Digital communication has proved itself to be more reliable and to have advantages over the analogue communication system. The achievements made by digital communication are robust and reliable. However, this field suffers from a major problem at the receiving end, which is known as Inter Symbol Interference (ISI) due to which one symbol overlaps with the subsequent symbols.

A very common approach to overcome this problem is by using Decision Feedback Equalizer (DFE). The filters used here are adaptive filters where the coefficients get updated with the help of Least Mean Square (LMS) algorithm. The LMS algorithm is convenient due to its computational simplicity. However, it has very low convergence speed.

The main aim of this paper is to improve the existing algorithm in a way that would converge faster and produce a better mean square error. This is done by implementing some constrains in the filter coefficients updating criteria. According to simulation results, the improvements provide fast convergence of the algorithm and low mean square error.

## II. REVIEW OF THE STATE OF ART

One of the first solutions to the elimination of ISI was to build a linear equalizer which mainly consists of linear filter and a threshold device. To optimize the transmitter and receiver, the mean square error between the input and the output of the equalizer was minimized. To simplify the mathematical problem, zero forcing conditions and mean square error were used extensively. Basically the approach was to minimize noise variance using zero forcing constraint, minimize the mean square error and also minimize the probability of error [2].

Nonlinear equalizers have also been used and have proven to have better performance over the simple linear types. DFE is a non-linear equalizer. The feedback filter uses linear combination of previous outputs and uses them to minimize the mean square error. For simplification, it is assumed that the outputs which are used in the feedback part are correct [2] [1].

An improvement to DFE was made by using BLMS (Block implementation of LMS) where a block adaptive filter was used instead of the sample by sample DFE. A problem in DFE is the huge computation complexity which is due to the long feedback part of the DFE. In this solution the signal is divided into blocks. To get small delays, the length of the block can be much smaller than the filter's order. The block LMS based DFE is mathematically equal to the sample by sample version and at the same time offers a significant computational saving [3].

## III. PROBLEM STATEMENT AND MAIN CONTRIBUTION

The question here is, if LMS based DFE can be improved in a way to have fast convergence? We hypothesize that by modifying the filter coefficient updating procedure we can improve the existing LMS based DFE and get a faster convergence.

The main contribution is to model the adaptive filter with feedback equalizer and LMS algorithm and then to implement the model in Matlab. We validate the system by comparing the error and speed of convergence of our improved method with that of the conventional LMS method.

## IV. PROBLEM SOLUTION

### A. Modeling

We model the system according to DFE module presented in Fig. 1. The equalizer consists of FF (Feed Forward) filter, FB (Feed Back) filter and a detector which includes a quantizer and a decision device.

The FF and FB filters can be linear FIR (Finite Impulse Response) filters. The FB filter subtracts the previous transmitted symbols which are known from FF filter's output with proper weighting, leaving only the current transmitted symbol in the output. The nonlinearity of DFE originates form the decision device which supplies the input to the FB filter [1].

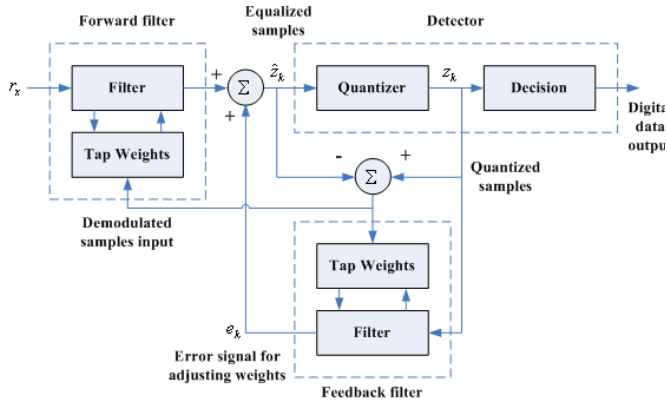

Fig. 1. Module of decision feedback equalizer

The weight coefficients for FF and FB filters are updated using LMS algorithm. The mean square error, formed from the subtraction of the output and the input of the quantizer, is used to update filter coefficients. The aim is to minimize this error. Once this goal has reached, FB filter provides better estimate of the ISI signal components and a clearer output is produced by DFE.

In order to overcome the slow convergence of LMS algorithm, we have proposed constrains in the updating process of filter coefficients. In our approach the step size is multiplied by the absolute value of the subtraction of two latest sequential errors. As result, the convergence would be fast in the starting iterations due to huge error at the beginning of the sequence. As the number of iterations increases, the mean square error would decrease and this leads to slower convergence, until the minimum mean square error is reached.

To implement in Matlab, We first produce a BPSK transmitted signal. In the next step, the signal is corrupted by passing through an ISI channel. The equalizer receives the ISI affected signal as input. The quantizer replaces every positive value of the input signal to the detector with „+1" and every negative value with „-1". A very simple form of quantizer could be a sign function. The improved LMS algorithm is used to minimize the mean square error formed from subtraction of output and input of quantizer. Assuming that the detector has so far made correct decisions, the input to the FB filter is the previous transmitted symbols provided by the detector

*B. Validation*

We validate our results by presenting the speed of convergence of our solution compared to the existing one. This comparison is shown in figure 2.

The blue color curve represents the error for the actual LMS based DFE while the red curve represents the error for the improved method. As can be seen from the figure, the improved method converges much faster (up to 6.2 times faster) than the actual method.

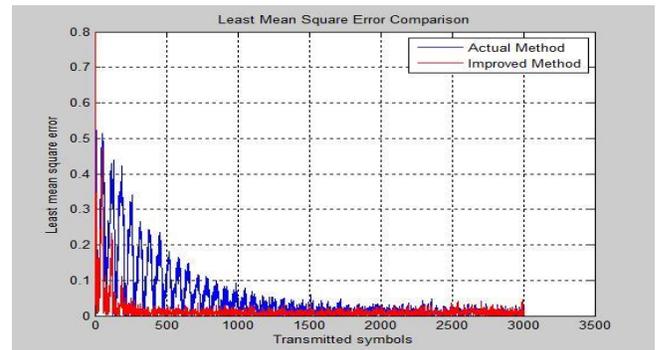

Fig. 2 Least mean square error comparison of the conventional method and the improved one

Furthermore, we can compare the least mean square error. from the input and output of the equalizer. As discussed before this error shows how close our estimates are to the real transmitted signal.

TABLE I
COMPARISON OF LEAST MEAN SQUARE ERROR OF ACTUAL METHOD AND IMPROVED METHOD

| Method | Least Mean Square Error |
| --- | --- |
| Conventional LMS | 5.6 |
| Improved LMS | 2.3 |

TABLE I shows the LMS error comparison between the two methods. As TABLE I shows, our improved method has less error compared to the actual existing one.

V. CONCLUSION

In this paper implementation of an improved LMS based DFE is carried out. Based on the results, we have shown that our improved method has higher convergence speed and less error in comparison with the existing LMS based DFE. Due to the improvements made, the new method can be useful in various applications specially those which include very long stream of transmitted data.

The study can be further extended by applying more constraints in step size or combining the current approach with the block implementation of LMS based DFE.

VI. ACKNOWLEDGMENT

We would like to thank Professor Wlodek Kuleza for his kind supervision, which helped us in successful completion of this project.

REFERENCES

[1] H. Hong and Taun do haug, "Decision Feedback Equalizer," on Nov 18, 2007, 6:25 am US/Central [online] Available: http://cnx.org/content/m15524/latest/.[Accessed:Sept.09, 2009]
[2] C.A. Belfiore and J.H. Park, "Decision feedback equalization", *IEEE International Conference*, Vol.67, Issue: 8, pp. 1143- 1156, 2005-06-28.
[3] K. Berberdis, T. Rontogiannis and S. Theodoridis, "Efficient block implementation of the LMS based DFE," *Proceedings 13th Digital Signal Processing International Conference*, Vol.1, pp. 143-146, July 1997

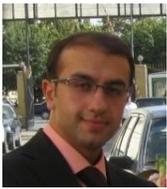
**Mohammad Havaei** was born in Isfahan, Iran in 1984. He holds a B.Tech in Electrical Engineering from Azad University of Najafabad Iran in 2006. Upon graduation, he worked as an engineering supervisor in Rahavard Institute. His area of interest includes Neural Networks, speech and image processing . He is currently undergoing his MSc study in Electrical Engineer with emphasis on Signal Processing at Blekinge Institute of Technology, Sweden.

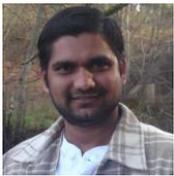
**Krishna Prasad Nandivada** was born in Hyderabad, India on April 24, 1984. He received his bachelor in Electronics and communication engineering from Jawaharlal Nehru Technological university, India. He is currently a student in Electrical Engineering at Blekinge Tekniska Hogskola (BTH).

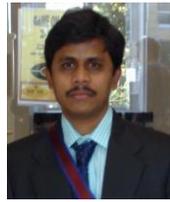
**Velleshala Sudheer** born in hyderabad, India in 1985. He completed his Bachelor degree in Electronics & Communication Engineering in 2006 from Jawaharlal Nehru Technological University, Hyderabad. He has worked as Testing Engineer in Hindustan Aeronautics Limited, Hyderabad from 2006 to 2008.
 He is currently pursuing his master"s in Electrical Engineering, with emphasis on Signal Processing at Blekinge Tekniska Hogskola, Karlskrona, Sweden. His area of interest is Radar engineering.